\documentclass{article}

\usepackage{arxiv}
\usepackage[square,sort,comma,numbers]{natbib}
\usepackage[utf8]{inputenc} 
\usepackage[T1]{fontenc}    
\usepackage{hyperref}       
\usepackage{url}            
\usepackage{booktabs}       
\usepackage{amsfonts}       
\usepackage{nicefrac}       
\usepackage{microtype}      
\usepackage{lipsum}		
\usepackage{graphicx}
\usepackage{natbib}
\usepackage{doi}
\usepackage{amsmath,amssymb}

\title{Semantic Video Moments Retrieval at Scale: A New Task and a Baseline}

\author{ Na Li \\
	Rutgers University\\
	Piscataway, NJ, USA \\
	\texttt{na.li@rutgers.edu} \\
	}
\date{}

\renewcommand{\headeright}{}
\renewcommand{\undertitle}{}
\renewcommand{\shorttitle}{}

\begin{document}
\maketitle

\begin{abstract}
	Motivated by the increasing need of saving search effort by obtaining relevant video clips instead of whole videos, we propose a new task, named \textbf{S}emantic \textbf{V}ideo \textbf{M}oments \textbf{R}etrieval at scale (SVMR), which aims at finding relevant videos coupled with re-localizing the video clips in them. Instead of a simple combination of video retrieval and video re-localization, our task is more challenging because of several essential aspects. In the 1st stage, our SVMR should take into account the fact that: 1) a positive candidate long video can contain plenty of irrelevant clips which are also semantically meaningful. 2) a long video can be positive to two totally different query clips if it contains clips relevant to two queries.
The 2nd re-localization stage also exhibits different assumptions from existing video re-localization tasks, which hold an assumption that the reference video must contain semantically similar segments corresponding to the query clip. Instead, in our scenario, the retrieved long video can be a false positive one due to the inaccuracy of the first stage. 
To address these challenges, we propose our two-stage baseline solution of candidate videos retrieval followed by a novel attention-based query-reference semantically alignment framework to re-localize target clips from candidate videos.
Furthermore, we build two more appropriate benchmark datasets from the off-the-shelf ActivityNet-1.3 \cite{caba2015activitynet} and HACS \cite{zhao2019hacs} for a thorough evaluation of SVMR models.
Extensive experiments are carried out to show that our solution outperforms several reference solutions.
\end{abstract}

\section{Introduction}

Video content generation is growing rapidly and video content analysis plays an essential role in understanding and processing such data. Content-based video retrieval \cite{chang1998fully,ren2009state} can enable to search for similar videos from a large collection of videos via a video query. Nevertheless, it is always the case people may only want the precisely matched part of the retrieved videos. Effectively re-localizing a query video clip from a large video gallery is of great potential to achieve such a goal.
Recently, video re-localization task is proposed \cite{dong2018video,feng2019spatio} to find semantically similar clips from a reference video according to a query clip. However, this task is limited to re-localization from a single video. Also, it holds a assumption that the reference video must contain semantically similar segments corresponding to a query clip. 
So it still remains an open question how to effectively and efficiently localize a query clip from a large video dataset and there is rare research into this direction.

\begin{figure}
\vspace{-1em}
\begin{minipage}[t]{\textwidth}
\begin{minipage}{0.6\textwidth}
\centering
\includegraphics[scale=0.35]{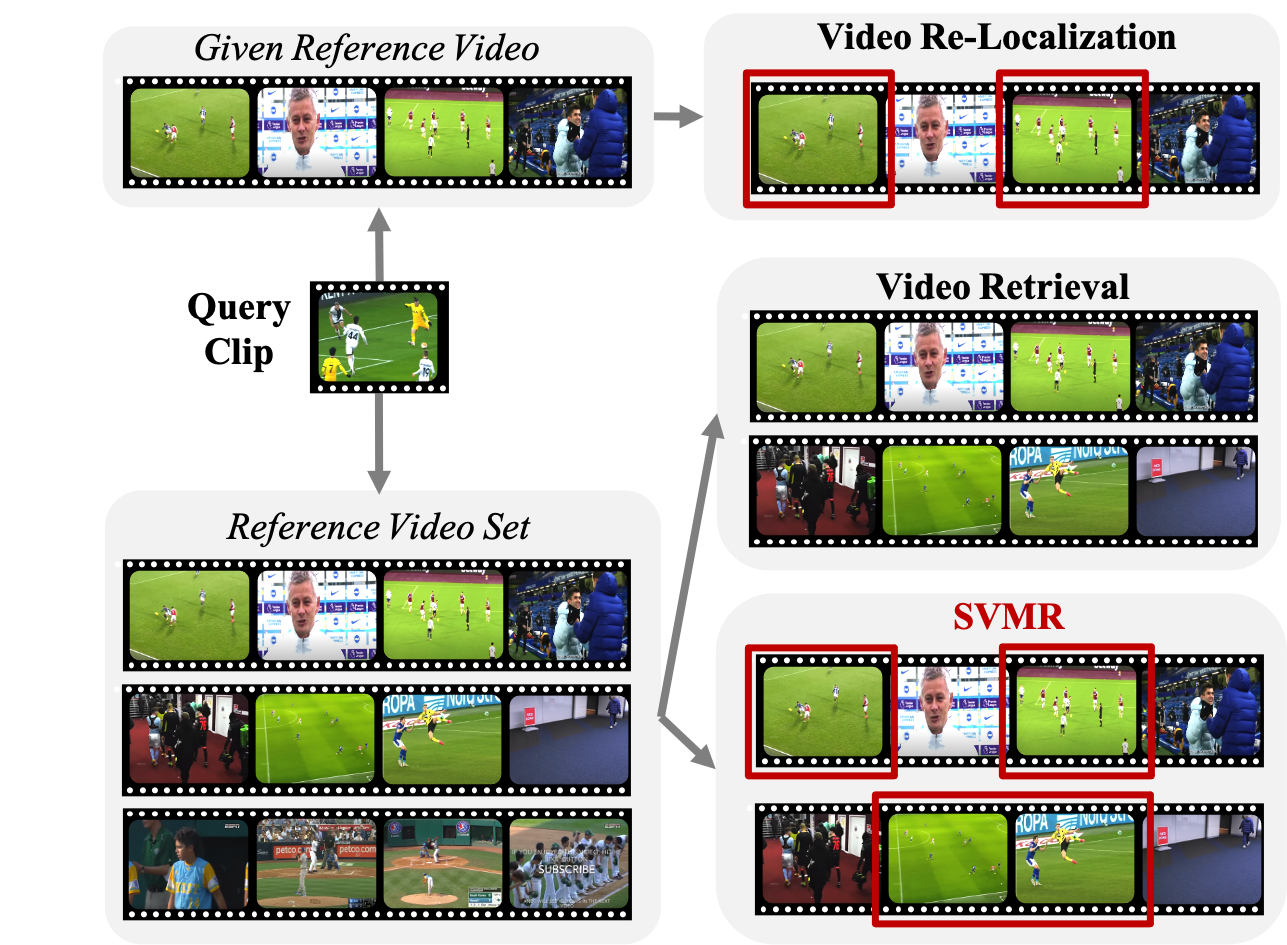}
   \caption{Illustration of problem settings of video retrieval, video re-localization and the proposed new SVMR task.} 
\label{illustration}
\end{minipage}
\hfill
\begin{minipage}[t]{0.4\textwidth}
\centering
\includegraphics[scale=0.16]{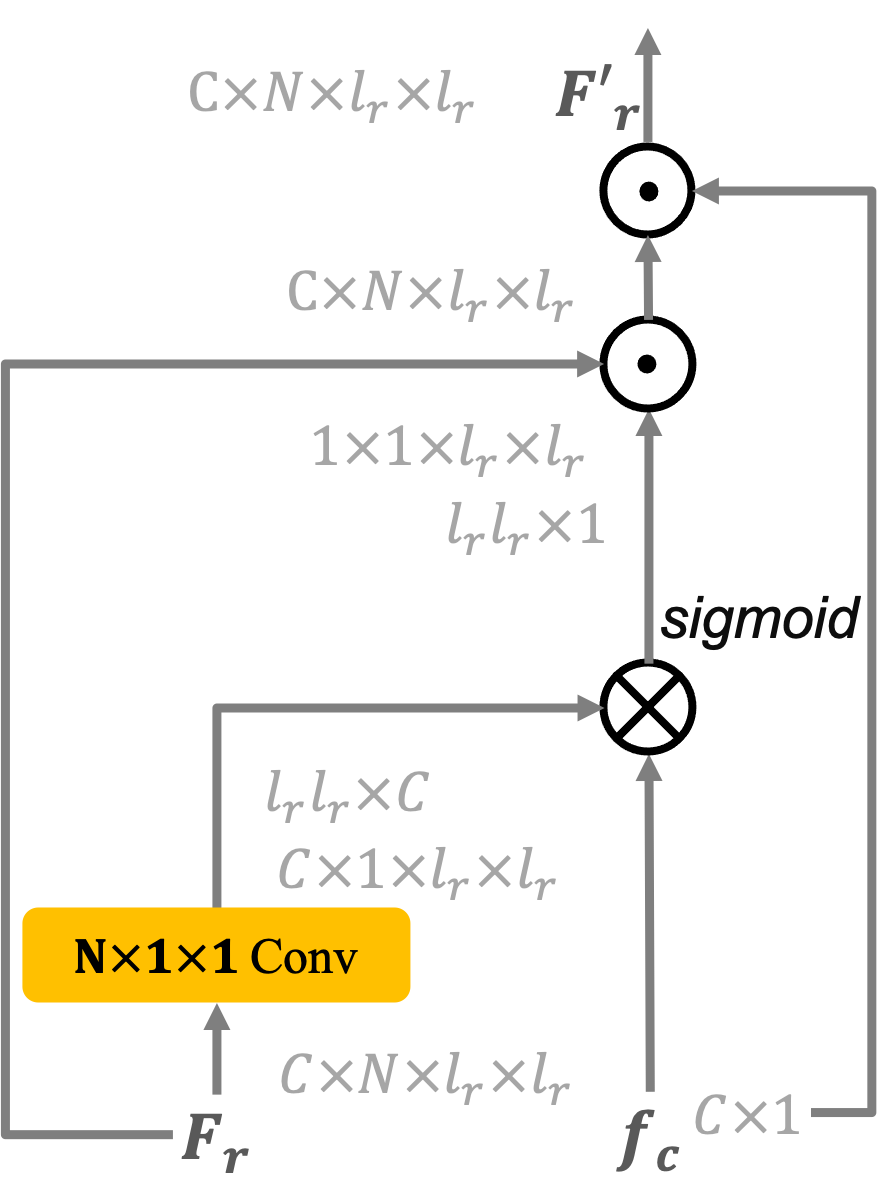}
\caption{Re-Localization Attention in ATLN.}
\label{fig:att}
\end{minipage}
\end{minipage}
\vspace{-1em}
\end{figure}

Therefore, in this work, we firstly bring up a new task named \textbf{S}emantic \textbf{V}ideo \textbf{M}oments \textbf{R}etrieval at scale (SVMR) to extend the existing research topics and to meet the real application requirement, as illustrated in Figure \ref{illustration}.

It is noticed that, instead of a simple combination of conventional video retrieval and video re-localization, it is more challenging because the task itself is much more complicated. In our 1st stage, we should consider that: 1) a positive candidate long video can contain plenty of irrelevant clips which are also semantically meaningful. 2) a long video can be positive to two totally different query clips if it contains clips relevant to the two queries. Our 2nd stage also exhibits different scenarios from existing video re-localization tasks. Conventional re-localization task is limited to re-localization from a single video. Also, it is limited by assumpting that the reference video must contain semantically similar segments corresponding to a query clip. 

In addition, feature correspondence modeling between query clips and video clips is non-trivial because of large intra-class variation among semantically similar video clips. Possibility of more than one target or an interferential clip inside one video further makes it harder. 
Then, a SVMR framework has to satisfy: 1) it is possible to pre-calculate features for the gallery to enable efficient search; 2) it can precisely localize semantic similar clips from the gallery. 

Moreover, we have noticed that there is no off-the-shelf dataset collected specifically for our task. Famous video retrieval dataset V3C1 \cite{berns2019v3c1} as well as video action localization datasets including ActivityNet-1.3 \cite{caba2015activitynet} and HACS \cite{zhao2019hacs} are all not directly applicable. 
In V3C1, the query is text rather than a video clip. 
ActivityNet-1.3 and HACS are widely used for temporal action localization \cite{buch2017sst,heilbron2016fast,lin2018bsn,lin2019bmn} with temporal boundaries available. However, a large number of videos contain only one action class (both over 99\% in ActivityNet-1.3 and HACS) or only one action instance (over 75\% in ActivityNet-1.3), thus it lacks diversity and will limit generalization of the learned model in scenarios where multiple targets or irrelevant clips exist in one video. 

To this end, we reprocess the off-the-shelf ActivityNet-1.3 \cite{caba2015activitynet} and HACS \cite{zhao2019hacs} datasets to form two benchmarks for comprehensively evaluating performance of SVMR.
Subsequently, we propose our two-stage solution of candidate reference videos retrieval followed by a novel attention-based query-reference semantic alignment framework. 
We design a two-branch auto-encoder framework, one for query encoding and the other for candidate encoding, for reference videos retrieval. We take into account the fact only part of a video may be a positive target in training at this stage. The query-reference semantic alignment framework can densely enumerate candidate clips from a reference video and model the correspondence between candidate clip-query pairs via a novel attention-based mechanism for effective re-localization. 
Extensive experiments are carried out on these benchmarks to show that our solution outperforms several reference solutions.

To summarize, our main contributions are as follows:
\begin{itemize}
    \item We firstly propose a new task termed \textbf{S}emantic \textbf{V}ideo \textbf{M}oments \textbf{R}etrieval at scale (SVMR), which aims at simultaneously retrieving videos and re-localizing semantic similar clips from retrieved videos via a query clip, to meet real application requirement;
    \item Two new benchmarks built from ActivityNet-1.3 \cite{caba2015activitynet} and HACS \cite{zhao2019hacs} are constructed for comprehensive evaluation of SVMR and they will be released to the community for further study in the future;
    \item A novel two-stage framework, which firstly obtains candidate reference videos via similarity search by leveraging a two-branch auto-encoder framework and then applies query-reference semantic re-localization via our attention-based alignment model, is proposed as a baseline for SVMR and extensive experiments are carried out to validate its effectiveness and superiority. 
\end{itemize}

\section{Related Work}
\subsection{Video Retrieval}
Content-based video retrieval \cite{chang1998fully,ren2009state,dong2018video} has been developed for decades. Recent deep learning-based solutions mainly focus on learning compact video representations for both recognition and retrieval, they can be categorized into supervised settings (such as C3D \cite{c3d} and \cite{p3d}) or unsupervised learning (\textit{e.g.}, RSPNet \cite{chen2020rspnet}, SpeedNet \cite{speednet}, \cite{lee2017unsupervised}, VCP \cite{vcp} and \cite{buchler2018improving}). These methods focused on either improving deep CNN backbones for video modeling or leveraging self-training for better big data exploration. Video retrieval has currently been advanced from content-based video retrieval to cross-modality settings, for example, text-video retrieval methods \cite{dong2021dual,song2019polysemous,wei2020universal} tried to map text and video embeddings into a common space for similarity measurement, \cite{suris2018cross} studied audio-video retrieval problem and multi-modality video retrieval, \cite{gabeur2020multi} leveraged transformers to learning multi-modal video features. These research works all share a common set of obtaining video-level retrieval results and are different from each other in their queries. They cannot localize the clip-of-interest from each video thus our SVMR task makes the first attempt to explore a more complex research problem.

\subsection{Moments Localization From Video}
Temporal action localization is a widely studied problem which aims at localizing a close-set of actions inside each input video. 
\begin{figure*}[tb]
\centering
\includegraphics[height=45mm,width=0.9\textwidth]{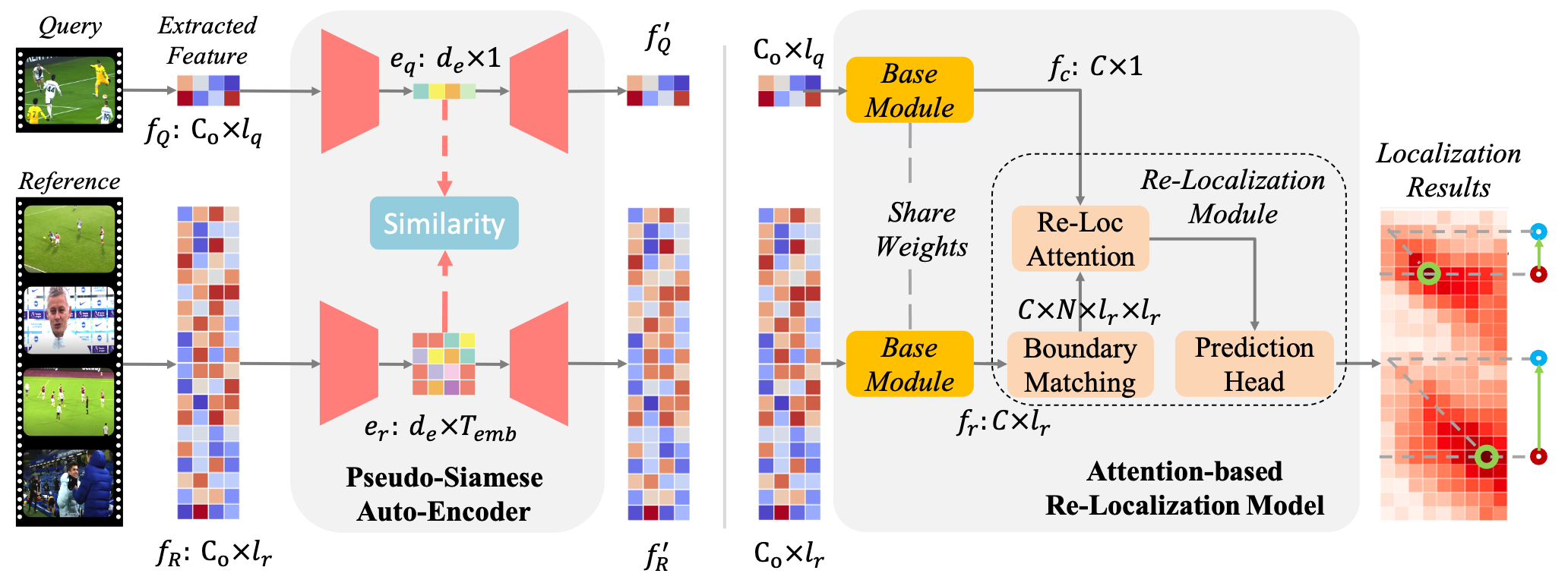}
\vspace{-0.5em}
\caption{Architecture of our proposed model for SVMR. It consists of a two-branch auto-encoder to retrieve relevant reference videos and an Attention-based Re-Localization Network to localize the corresponding segments from candidates.}
\label{fig:arch}
\vspace{-1.5em}
\end{figure*}

Most previous methods such as \cite{buch2017sst,heilbron2016fast,escorcia2016daps,gao2017turn,shou2016temporal} adopt a ``top-down'' fashion to generate action proposals from predefined temporal anchors. Some recent works, \textit{e.g.} SSN \cite{zhao2017temporal}, BSN \cite{lin2018bsn}, BMN \cite{lin2019bmn} and RapNet \cite{gao2020accurate}, propose ``bottom-up'' action proposal generation framework with more precise boundaries and more flexible action duration. Temporal action localization can obtain clip-granularity segments from videos, but they are restricted to some closed-set action labels. Another hot research topic in terms of moments retrieval in video called video grounding is introduced in TALL \cite{gao2017tall} and MCN \cite{anne2017localizing}. It can retrieve moments from a video by natural language queries.
In 2D-TAN \cite{zhang2020learning}, it leverages the state-of-the-art top-down action localization framework BMN \cite{lin2019bmn} and adds a natural language embedding fusion module for video grounding and achieves performance improvement. 
Reinforcement learning-based frameworks \cite{he2019read,wang2019language,wu2020tree} are also proposed to treat this task as a sequential decision-making problem. 
Video re-localization \cite{feng2018video} is introduced by Feng \textit{et al.}. 
In this task, a model is required to localize a semantic similar clip from a reference video according to an input query video clip. Later, spatial-temporal video re-localization is studied in \cite{feng2019spatio}. Recently, there are works \cite{sun2021vsrnet,hou2021conquer} address the problem of retrieving a video clip from a large video gallery, but they target text-to-clip retrieval.

\section{The Task and Benchmarks}
\subsection{Problem Formulation}

As shown in Figure \ref{illustration}, given a query clip $v_q$ and a reference untrimmed video set $R = \{v_{r_i}\}_{i=1}^{N_r}$ with $N_r$ videos, the goal of SVMR task is two-fold:
Firstly, we need to search from $R$ to find the top $N_c$ candidate videos containing clip(s) semantically similar to $v_q$, which we denote as candidate set $G = \{v_{c_i}, p_{v_q,c_i}\}_{i=1}^{N_c}$, where $p_{v_q,c_i}$ is the similarity score.
Secondly, for each candidate untrimmed video $v_{c_i}$, we need to temporally localize timestamps of semantically similar clip(s) of $v_q$, which are denoted as $\Psi_{c_i} = \{\varphi_{m, c_i}=(t_{s,c_i}^{(m)},t_{e,c_i}^{(m)}, s_{c_i}^{(m)})\}_{m=1}^{M_c}$, where $t_{s,c_i}^{(m)},t_{e,c_i}^{(m)}$ and $ s_{c_i}^{(m)}$ are starting time, ending time and confidence score of $m^{th}$ instance $\varphi_{m, c_i}$, respectively.
Thus, given $v_q$ and $R$, the final retrieved results can be denoted as $\{\Psi_{c_i}\}_{i=1}^{N_c}$. For example, given a query clip of ``brushing teeth'', the candidate videos are those containing the same action and the action clips inside those candidates are final SVMR targets.

\subsection{Dataset Construction}
To overcome shortcomings of lacking diversity in the original ActivityNet-1.3 \cite{caba2015activitynet} and HACS \cite{zhao2019hacs}, we re-organize them to form two new benchmarks for SVMR. Taking HACS for example, we first merge its train and validation videos and randomly split them into two even parts $V_1$ and $V_2$. All the action clips inside $V_1$ are extracted to serve as queries. There are 200 action classes in HACS in total. Among the extracted clips, the ones that belong to the first 160 action classes are used as training queries, the ones that belong to the next 20 action classes form the validation queries and the clips of the rest 20 classes are used for testing queries. $V_2$ is used to generate the reference set as follows. 1/3 of videos in $V_2$ is kept unchanged. To construct videos with no less than 2 actions, firstly we randomly sample a base video $v_b$ and a supplementary video $v_s$ such that they have no common action labels from the rest of 2/3 videos. Then we randomly sample several segments from $v_s$ such that each contains an action instance and randomly insert them to $v_b$ at a random background timestamp. This procedure is repeated by $1/3\times|V_2|$ times. In Analogy, we can construct videos with no less than 3 actions by randomly sampling 1 base video and 2 supplementary videos and then randomly merging them. 

\section{Methodology}

In this section, the proposed two-stage baseline framework for the proposed SVMR task is described in detail. It is composed of (1) a two-branch auto-encoder for video-level ranking and (2) an Attention-based Re-Localization Network (ARLN) for localizing the corresponding segments in candidate videos. Please note we follow convention practices \cite{lin2018bsn,lin2019bmn,feng2018video} and build our models upon visual feature sequence extracted from raw videos. 
For query $v_q$ and a reference video $v_r$, we denote their extracted feature sequence as $f_Q \in R^{C_o \times l_q}$ and $f_R\in R^{C_o \times l_r}$. Here $C_o$ is the dimension of feature vector,  $l_q$ and $l_r$ are the length of $f_Q$ and $f_R$.
Figure \ref{fig:arch} is a block diagram of our framework. The left part illustrates our two-branch auto-encoder model and the right part shows the ARL model.

\textbf{In the first stage}, fix-size feature embedding for query and reference videos are generated separately. The query feature $f_Q$ is embedded into $e_q \in R^{d_e}$, while the reference feature $f_R$ is embedded into $e_r \in R^{d_e\times T_{emb}}$. Similarity between query and reference video is then calculated as their maximum cosine similarity:

\begin{equation}
p_{q,r} = \underset{i=1}{\overset{T_{emb}}{max}} \frac{e_q \cdot e_r[i]}{\left \|e_q  \right \|\cdot \left \|e_r[i]  \right \| }
\end{equation}

Based on similarity score, we can ranking all reference videos and get top $N_c$ candidate videos $G = \{v_{c_i}\}_{i=1}^{N_c}$.
\textbf{In the second stage}, given query and ranked candidate videos, we propose a \emph{Attention-based Temporal Localization Network (ATLN)} for locating the corresponding segments in each candidate videos, and generate final retrieved results $\{\Psi_{c_i}\}_{i=1}^{N_c}$.

\subsection{Two-Branch Auto-Encoder}

To achieve video-level retrieval, we need to generate a compact and distinguishable feature embedding for both query and references videos. There are two main difficulties: (1) reference video may be much longer than query clip; (2) complex and confusing background content exist in the untrimmed reference video.
To address these difficulties, as shown in Figure \ref{fig:arch},  we proposed a two-branch auto-encoder framework to extract embedding for query and references separately.

\textbf{Encoder:}
We design two asymmetric branches with independent weights to deal with varying temporal lengths between the query clip and the whole reference video, which is inspired by Siamese architectures \cite{chopra2005learning}\cite{koch2015siamese} for comparing the similarity of two inputs.

Specifically, the query clip $f_Q$ goes through the query sub-network and is embedded into  $e_q \in R^{d_e}$, and a reference video $f_R$ goes through the reference sub-network and is embedded into $e_r \in R^{d_e\times T_{emb}}$.

Both sub-networks contain three convolution layers, while different number of average pooling layers are used such that the query embedding and reference embedding are with length of $1$ and $T_{emb}$, respectively. To achieve more representative embeddings, inspired by previous wisdom \cite{li2020deep,tgm}, we adopt the Concept-wise Temporal Convolution (CTC) layer \cite{li2020deep} in our two-branch auto-encoder instead of the vanilla temporal 1d convolution layer. In CTC layer, given a feature tensor with shape of $C_{in}\times L\times F_{in}$ ($f_Q$ and $f_R$ can be viewed as $F_{in}$ equals 1, here $L$ denotes temporal length of feature sequence and $C_{in}$ denotes input feature channels), each channel of the tensor is regarded as a concept, and a series of $F_{out}$ learnable temporal $1\times3$ filters are applied to each concept separately to produce output tensor shaped as $C_{in}\times L\times F_{out}$. With this setting, common temporal patterns of different concepts can be captured, which is helpful for generating discriminative feature embeddings. In our implementation, $F_{out}$s of the three CTC layers are empirically set to 1, 32 and 1, respectively.

\textbf{Decoder:}
The architecture of decoders can be seen as a reverse of encoders. Both decoders contain three convolution layers but the two branches contain different number of upsampling layers accordingly to the number of downsampling layers in the decoder, such that the reconstructed query feature $f_Q^{'}$ and reference feature $f_R^{'}$ share the same shape of their original input feature tensors. In our decoders, CTC layers are also leveraged. 
The constructed auto-encoders are trained by minimizing the reconstruction error to make sure the learned latent feature is meaningful. 

\subsection{Attention-based Re-Localization Network}

After ranking all reference videos and get top $N_c$ candidate videos, we need to further accurately locate the corresponding segments related to query clip in each candidate video.
To achieve this, we revise the state-of-the-art temporal localization method BMN \cite{lin2019bmn} as \textbf{A}ttention-based \textbf{R}e-\textbf{L}ocalization \textbf{N}etwork (ARLN), where a Re-localization Attention module is proposed for automatically evaluating similarity between query and reference temporally.

The original BMN mainly contains three modules, we revise it to suit video re-localization task as following:
(1) \emph{\textbf{Base Module}}: 

We adopt base module for both query and reference video with shared temporal 1D convolution layers. An average-pooling layer is followed after query base module to reduce temporal dimension to 1. Namely, for the input $f_Q\in R^{C_o\times l_q}$ and $f_R\in R^{C_o\times l_r}$, the base module will encode them to be $f_c$ with shape of $C\times 1$ and $f_r$ with shape of $C\times l_r$ feature tensors, respectively.
(2) \emph{\textbf{Proposal Evaluation Module (PEM)}} is revised to be \emph{\textbf{Re-Localization Module}} with attention-based query-proposal correspondence modeling: PEM is originally designed to generate Boundary-Matching (BM) feature map and evaluate confidence scores of densely-distributed temporal clips via generating BM confidence map. It is the key component of BMN. For readers who are unfamiliar with BMN, the Boundary-Matching can be simply regarded as a module which samples $N$ features from $f_r$ to represent each candidate temporal clip. The BM feature map enumerates all densely distributed temporal clips whose start timestamps ranges from $0$ to $l_r-1$ and whose duration ranges from $1$ to $l_r$, in total there are $l_r*l_r/2$ candidates. Therefore, after Boundary-Matching module, the BM feature map is with shape of $C\times N \times S \times D$ ($S$ and $D$ denote the start and duration axis respectively, and both equals $l_r$.) and only the upper-left triangle is valid. Please refer to BMN \cite{lin2019bmn} for more details.
In this paper, We revise it to be a Re-Localization Module (RLM) which can capture query-candidate correspondence with attention-based \cite{vaswani2017attention} correlation operations.

\noindent
\textbf{Re-Localization Module.}
The goal of Re-Localization Module (RLM) is to generate a confidence map, where each value represents the confidence score that measures how the corresponding temporal proposal is related to a query clip. 

As illustrated in Fig \ref{fig:att}, firstly we use the Boundary-Matching layer \cite{lin2019bmn} to sample from reference feature $f_r$ and generate BM feature map $F_r \in R^{C\times N\times l_r\times l_r}$, where $N$ is the number of sampled feature per candidate clip, the third and fourth dimensions are the clip start timestamp and its duration. So each $[0:C, 0:N, s, d]$ tensor from the upper-left triangle ($D<l_r-s$) of $F_r$ is the feature of a valid clip starts at $s$ and lasts for $d$ from $f_r$.
Then, we introduce the Re-Localization Attention, which combines query and reference feature twice from different aspects.
\textbf{First}, $F_r$ is downsampled via a $N\times1\times1$ convolution to a $C\times1\times l_r \times l_r$ tensor and then it is reshape to be $F_r^{(d)}\in l_rl_r\times C$. Thus we can calculate attention scores as:

\begin{equation}
    Attention(f_c,F_r^{(d)}) = sigmoid(F_r^{(d)} \times f_c).
\end{equation}

The attention score is reshape to $Att$ with shape of $1\times1\times l_r\times l_r$ and is used to weight $F_r$, we can obtain:

\begin{equation}
    F_r^{(w)} = Att\cdot F_r
\end{equation}
\textbf{Second}, we apply Hadamard product to fuse query feature with the weighted reference feature to further exploit the intersection information between the query and the semantically corresponding segments:

\begin{equation}
    F_r^{'} = F_r^{(w)}\odot f_c
\end{equation}
Finally, following \cite{lin2019bmn}, we use two 2D convolution layers with sigmoid activation to predict a Boundary-Matching classification score map $M_c\in R^{l_r\times l_r}$ and a Boundary-Matching Regression score map $M_R\in R^{l_r\times l_r}$. The upper-left triangular parts of these score maps are valid and they describe the semantical similarity of the query and the densely-distributed candidate temporal clips.

\subsection{Training}
In this work, the retrieval model and re-localization model are trained separately. The training details of two stages are introduced in this section.

\subsubsection{Loss of Two-branch Auto-Encoder}

To achieve better embedding quality, we train the Pseudo-Siamese Auto-Encoder with re-construction loss and similarity loss jointly. 

\noindent
\textbf{Training Data Construction}
To train the stage1 model, at each iteration, we randomly sample a query clip $f_Q$ from the query set. Then, we randomly pick reference video $f_R$, with 50$\%$ possibility containing positive clip(s).
After picking reference video, we can generate the corresponding label $g \in R^{T_{emb}}$. Here, $g_i = 1 $ if the corresponding clip $[\frac{i}{T_{emb}}\cdot l_r, \frac{i+1}{T_{emb}}\cdot l_r]$ is semantical the same as the query clip, otherwise $g_i$ equals -1.

\noindent
\textbf{Re-construction Loss.}
The re-construction loss is applied to both branches, forcing the auto-encoder to accurately reconstruct input features and keep compact feature embedding. 
Denoting the input feature as $f$ and the reconstructed feature as $f'$, re-construction loss is obtained by:

\begin{equation}
    L_{recon}(f, f') = {\left \| f- f' \right \|_2}
\end{equation}

\noindent
\textbf{Similarity Loss.}
Moreover, we have the similarity distance loss between query clip and reference video as:

\begin{equation}
    L_{sim} = \frac{1}{T_{emb}}\sum_{i=1}^{T_{emb}}\left \| \frac{e_q\cdot e_{r}[i]}{\left \| e_q \right \| \cdot \left \| e_{r}[i] \right \|} -g_i \right \|_2,
\end{equation}

The overall loss of stage 1 can be summarized as follows:

\begin{equation}
    L_{1} = L_{recon}(f_Q,f'_Q) + L_{recon}(f_R,f'_R)  + \lambda \cdot L_{sim}
\end{equation}
where $\lambda $ is set as 2 to balance the loss terms.

\subsubsection{Loss of ATLN}
To achieve accurate video re-localization, we follow BMN \cite{lin2019bmn} to train the ATLN model with temporal evaluation loss and proposal evaluation loss jointly.

\noindent
\textbf{Training Data Construction}
To train the ATLN model, at each iteration, we randomly sample a query clip $f_Q$ and a reference video $f_R$ containing positive clip(s) from reference set.
Some reference videos may contain instances of multiple action categories, here we only consider those have the same category of the query clip as positive and regard other instances as background.

Following the proposal evaluation loss in \cite{lin2019bmn}, we set the loss of re-localization module as:

\begin{equation}
    L_{RLM} = L_C(M_C, G_C) + \lambda \cdot L_R(M_R, G_C)
\end{equation}
where $M_C$, $M_R$ are BM confidence maps and $G_C$ is the groundtruth 2D label map. $L_C$ is classification loss which can be calculated as binary logistic regression and $L_R$ represents $L_2$ regression loss.

\begin{table*}[!t]
\centering
\scalebox{0.83}{
\begin{tabular}{l|cccc|ccccc}
\hline
Config. &HR@1 &HR@5 &HR@10 &HR@50 &mAP@1 &mAP@5 &mAP@10 &mAP@50 &mAP@100 \\ 
\hline
Conv1D w/ AE &0.8680 &0.9549 &0.9722 &0.9933 &0.8680 &0.8906 &0.8781 &0.8314 &0.7993 \\ 

CTC\_layer &0.8646 &0.9509 &0.9697 &0.9903 &0.8646 &0.8901 &0.8792 &0.8356 &0.8068 \\ 

CTC w/o AE &0.8713	&\textbf{0.9645}	&0.9665	&0.9955 &0.8713	&0.8992	&0.8866	&0.8426	&0.8158 \\ 

CTC w/ AE Shared &0.7870 &0.9182 &0.9488 &0.9880 &0.7870 &0.8270 &0.8140 &0.7625 &0.7304 \\

CTC w/ AE  &\textbf{0.8835} &0.9604 &\textbf{0.9744} &\textbf{0.9947} &\textbf{0.8835}	&\textbf{0.9060}	&\textbf{0.8951}	&\textbf{0.8498}	&\textbf{0.8191} \\

\hline
\end{tabular}}
\caption{Analysis on effectiveness of each component of our two-branch auto-encoder. ``CTC'' and ``AE'' stand for CTC\_layer and auto-encoder reconstruction loss, respectively.}
\label{stage1}
\end{table*}

\section{Experiments}
\textbf{Datasets and Evaluation Metrics}
Our experiments are carried out on the two constructed benchmarks based on ActivityNet-1.3 and HACS. In the following, without extra specifying, we directly name the two re-organized datasets as ANET and HACS for convenience. Note that the clips used for training, validation and testing are with different action labels, respectively. 
such a setting can be more suitable to evaluate the generalization ability of clip-granularity semantic video retrieval models. 
To evaluate the performance of stage 1, we use two widely adopted metrics of top $K$ hit rate (HR@K) and top $K$ mean average precision (mAP@K). 
Please note that a retrieved clip is considered positive when its temporal IoU (tIoU) with a groundtruth positive sample is greater than a threshold $\tau$. At the second stage, following \cite{lin2018bsn,lin2019bmn}, we calculate the Average Recall (AR) under multiple tIoU thresholds $[0.5 : 0.9 : 0.1]$. We calculate AR under different Average Number of Proposals(AN) as $AR@AN$ and the Area under the AR v.s. AN curve (AUC) is adopted as evaluation metric. For end-to-end performance evaluation, the Prec@$N$ retrieved clips is reported.

\textbf{Implementation Details}
We use ResNet50 I3D\cite{carreira2017quo}, SEResNeXt-152 TSN\cite{wang2016temporal} and SEResNeXt-101 TSM\cite{lin2019tsm} to extract feature sequences from the original videos. The I3D, TSN and TSM models are pretrained on Kinetics600 \cite{carreira2017quo}.
At the first stage, input query features and reference features are resized to meet $l_q$=4 and $l_r$=100 by linear interpolation, respectively. $d_e$ equals 512 and $T_emb$ is empirically set to 4. $C=128$ and $N$ is 32 at the second stage.  In the training phase, both stages are trained independently. Adam optimizer with a learning rate of 0.0001 is adopted and batch size is set to be 256. The models are trained to converge using the training query set and are validated on the validation query set. In the inference phase, we report evaluation results using the testing query set. Top 10 candidate reference videos are obtained from stage 1 and the final retrieved clips are localized from them. 
In stage2, Gaussian weighted soft-NMS with sigma of 0.4 is applied. 

For a query $q$ and a clip from video $r$ which starts at $s$ and lasts for $d$, the final confidence score is $p_{q,r}*M_R[s,d]*M_C[s,d]$, where $p_{q,r}$ is their maximum cosine similarity obtained at stage 1 and $M_R$ and $M_C$ are boundary matching confidence maps obtained at stage 2.

\begin{minipage}{\textwidth}
\begin{minipage}[t]{0.45\textwidth}
\centering
\makeatletter\def\@captype{table}
\scalebox{0.8}{
\begin{tabular}{l|ccccc}
\hline
$T_{emb}$ &mAP@1 &mAP@5 &mAP@10 &mAP@50\\ 
\hline
2   &0.8611 &0.8915 &0.8765 &0.8229 \\
4   &\textbf{0.8835} &\textbf{0.9060} &\textbf{0.8951} &0.8498\\
8   &0.8794 &0.9028 &0.8933 &\textbf{0.8558}\\ 
\hline
\end{tabular}}
\caption{Empirical analysis on impact of $T_{emb}$. }
\label{temb}
\end{minipage}
\begin{minipage}[t]{0.45\textwidth}

\centering
\makeatletter\def\@captype{table}
\scalebox{0.8}{
\begin{tabular}{l|ccccc}
\hline
&mAP@1 &mAP@5 &mAP@10 &mAP@50\\ 
\hline
Triplet &0.8636 &0.8895 &0.8748 &0.8136  \\ 
ArcFace  &0.7409 &0.7669 &0.7522 &0.6859\\ 
\hline
Ours  &\textbf{0.8835} &\textbf{0.9060} &\textbf{0.8951} &\textbf{0.8498}\\
\hline
\end{tabular}}
\caption{Two-branch v.s. One-branch.}
\label{table_4}
\end{minipage}
\end{minipage}

\subsection{Verification of Design Choices}
In this subsection, we provide experimental results to validate our design choices in both stages. We conduct experiments using HACS dataset considering its larger data volume than ANET. TSM \cite{lin2019tsm} based video features are used for our evaluation in this section and tIoU threshold $\tau$ is set to be 0.5.

\subsubsection{Ablation Study on Stage 1}
\paragraph{}
\textbf{Effectiveness of Each Component}
To verify effectiveness of our design choice in this stage, we carried out ablation analysis by disabling one model configuration at one time, and the same training policy is applied to these variants. The experimental results are summarized in Table \ref{stage1}. 
From the first and the last rows, we can see replacing CTC\_layer with conventional temporal 1D convolution layer will degrade the performance a lot, mAP@1 drops by 1.5\%. It shows the superiority of CTC\_layer over temporal 1D convolution. We also check that with the two branches sharing their weights, the performance drop is quite significant. up to 9.6\% in mAP@1. This is because the model capacity is reduced with shared weights meanwhile the query and reference data distributions may be different. From this point of view, no weight-sharing is quite important, as is our current implementation choice. When no auto-encoder reconstruction loss is applied, the mAP@1 will also drop from 88.35\% to 87.13\%, showing that auto-encoder configuration also makes a difference. 

\textbf{Impact of $T_{emb}$} Considering the existence of multiple action instances in one video, we carefully design the reference auto-encoder branch to encode reference videos to a $d_e\times T_{emb}$ hidden space and apply max cosine similarity loss to train the model. Intuitively, smaller $T_{emb}$ will lose temporal resolution and can make short and fast actions be swallowed by background in the hidden feature space, meanwhile larger $T_{emb}$ results in larger feature storage cost of the gallery and makes the search process slower. In order to analyze the impact of hyper-parameter $T_{emb}$, we empirically validate it via experiments by varying the $T_{emb}$ while keeping other settings unchanged. The
results are summarized in Table \ref{temb}. Empirically, $T_{emb}$ equals 4 achieves the best performance at the first candidate video retrieval phase. 

\textbf{Two-branch v.s. One-branch} At stage 1, our model works in a two-branch fashion, which is quite different from existing content-based video retrieval methods. Conventional retrieval methods typically adopt a single model to map both query videos and reference videos into a common hidden space. In order to show the superiority of our two-branch design, we compare our stage 1 model to single-branch model trained with popular ranking loss functions, \textit{i.e.}, triplet loss \cite{hoffer2015deep} and ArcFace \cite{deng2019arcface}. The results are shown in Table \ref{table_4}.
We can observe that our two-branch fashion achieves better performance than conventional one-branch fashion when it comes to the new SVMR task. In this task, the query may be only similar with a specific part of the reference videos, one-branch solutions have to map both query and reference into a common feature space, which can cause feature aliasing for a long reference video and degrades the retrieval performance. Note that ArcFace drops greatly, this is because it requires groundtruth action labels for training, in our setting, a reference video may contain multiple action classes and then its label must be randomly chosen from these classes. This process will greatly confuse the network.

\begin{table*}[!t]
\centering
\scalebox{0.9}{
\begin{tabular}{l|cccc|ccccc}
\hline
Config. &HR@1 &HR@5 &HR@10 &HR@50 &mAP@1 &mAP@5 &mAP@10 &mAP@50 &mAP@100 \\ 
\hline

HACS TSM &0.8835 &0.9604 &0.9744 &0.9947 &0.8835	&0.9060	&0.8951	&0.8498	&0.8191 \\

HACS I3D &0.7520 &0.8924 &0.9285 &0.9785 &0.7520 &0.7873 &0.7654 &0.6795 &0.6293 \\ 

HACS TSN &0.8039 &0.9082 &0.9374 &0.9743 &0.8039 &0.8362 &0.8283 &0.7974 &0.7785\\  \hline

ANET TSM &0.7186 &0.8628 &0.9106 &0.9707 &0.7186 &0.7579	&0.7421	&0.6659	&0.6206\\
ANET I3D &0.7661 &0.8945 &0.9252 &0.9807 &0.7661 &0.8056 &0.7902 &0.7121 &0.6633\\
ANET TSN &0.8614 &0.9361 &0.9528 &0.9843 &0.8614 &0.8854 &0.8738 &0.8195 &0.7748\\

\hline
\end{tabular}}
\caption{Stage 1 retrieval performance using different features.}
\label{stage1_feats}
\end{table*}

\subsubsection{Ablation Study on Stage 2}
In this part, we evaluate localization performance under the assumption that the retrieved reference videos are positive.
The model is trained and tested with positive query-video pairs.

\paragraph{}
\textbf{Analysis on the attention mechanism}
Table\ref{stage2_att} evaluates the effectiveness of our attention mechanism and the query filter product operation. As we can see, both attention mechanism and query filter product operation are helpful for the SVMR task. Attention part helps our model to focus on those candidate proposals that are more related to query clips, with attention, AUC can be boosted from 59.37\% to 61.43\%. Moreover, a query filter product further exploits the inter-connections between query clips and candidate proposals, it significantly improves AUC from 54.90\% to 61.43\%. We also conduct experiments on how to integrate the query clips and weighted proposals. The results show that product operation achieves better performance than concatenation operation (61.43\% v.s. 59.24\% in AUC).

\begin{minipage}[t]{\textwidth}
\begin{minipage}{0.45\textwidth}
\centering
\makeatletter\def\@captype{table}
\scalebox{0.75}{
\begin{tabular}{l|ccc|r}
\hline
Method  &AR@5 &AR@10 &AR@100 &AUC \\ \hline
Attn+Concat &0.4000 &0.4800 &0.6734 &59.24\\ 
Prod.  &0.3951 &0.4690 &0.6833 &59.37 \\ 
Attn.  &0.3117 &0.4018 &0.6574 &54.90 \\ 
Attn+Prod. &\textbf{0.4137} &\textbf{0.4903} &\textbf{0.6984} &\textbf{61.43}\\ 
\hline
\end{tabular}}
\caption{Effect of each component of action re-localization module. ``Prod'' means directly product $F_r$ and $f_c$, ``Attn+Prod'' denotes our scheme (Fig. \ref{fig:att}). ``Attn'' and ``Attn+Concat'' mean removing the final production operation in Fig. \ref{fig:att}  and replace it by concatenation, respectively.}
\label{stage2_att}
\end{minipage}
\begin{minipage}{0.45\textwidth}
\centering
\makeatletter\def\@captype{table}
\scalebox{0.75}{
\begin{tabular}{l|c|ccc|r}
\hline
       &\# class  &AR@1 &AR@10 &AR@100 &AUC \\ \hline
w/o & 1 &0.1563 &0.3677	&0.6663	&53.56 \\
query  &2  &0.0768	&0.3541	&0.6512	&52.62 \\ 
branch  &3  &0.0487	&0.3467	&0.6296	&51.08 \\ \hline 
w/     &1  &0.2047 &0.4251 &0.6955 &58.79 \\
query  &2 &0.2439	&0.4986	&0.7109	&62.76\\
branch  &3 &0.2338	&0.5150	&0.6941	&61.89\\
\hline
\end{tabular}}
\caption{Performance evaluated on different subsets which contains videos with 1, 2 and 3 action classes, respectively.}
\label{table_classes}
\end{minipage}
\end{minipage}

\textbf{Why New Benchmarks are In Need} In this part, we compare experimental results of stage 2 of our solution (``w/ query branch'') with purely BMN \cite{lin2019bmn} (``w/o query branch'') considering videos with one, two, and three action classes inside, respectively. 
Results are presented in Table \ref{table_classes}. It shows the necessity of our newly constructed datasets. As we can see, the gap between ``without query branch'' and ``with query branch'' differs among the one, two, and three action classes. The gaps of two and three action classes are very large, meanwhile, when only one action class exists in reference videos the gap is much smaller. If ActivityNet-1.3 and HACS are not re-processed, most of their videos contain only one action class, the re-localization task of stage 2 can collapse to a temporal action proposal task \cite{lin2018bsn,ssad,lin2019bmn}. Therefore, such datasets cannot well benchmark re-localization performances, because query-proposal correlation would become trivial to the re-localization precision and the model only needs to distinguish foreground clips after all.

\subsection{Different Feature Extractors}
We evaluate how different feature extractors can affect SVMR performance and we report baseline results on both datasets. 
\textbf{stage1:}
In our video-level retrieval stage, based on results reported in Table\ref{stage1_feats}, we surprisingly find that different feature extractors behave differently on the two datasets. On HACS, the better the feature extractor is, the better the performance is. We can see the best performance is achieved by TSM \cite{lin2019tsm} among these feature extractors. However, in ANET, TSM achieves the worst performance and TSN achieves the best results among these feature extractors. 

This phenomenon may be caused by the difference in data distributions as well as dataset volumes. ANET is much smaller than HACS, therefore, a very strong feature is easier to suffer from overfitting issues for video-level retrieval given a clip-level query. 

\textbf{stage2:}
In the re-localization phase, baseline performances using different features are presented in Table 7.
TSM feature achieves the best results on both HACS and ANET, which is different from stage 1.  

\begin{minipage}[t]{\textwidth}
\begin{minipage}{0.5\textwidth}
\begin{minipage}{0.5\textwidth}
\makeatletter\def\@captype{table}
\centering
\scalebox{0.8}{
\begin{tabular}{l|ccc|r}
\hline
Config.  &AR@5 &AR@10 &AR@100 &AUC \\ \hline
HACS TSM  &0.4137 &0.4903 &0.6984 &61.43\\ 
HACS I3D &0.2963 &0.3676 &0.5890	&49.76 \\ 
HACS TSN &0.1829 &0.2309 &0.4481 &35.07\\  \hline
ANET TSM &0.4841 &0.5633 &0.7611 &67.45\\
ANET I3D &0.3760	&0.4490 &0.7164 &61.67\\
ANET TSN &0.4765 &0.5438 &0.7340	&64.96\\ \hline
\end{tabular}}
\parbox{5cm}{\caption{Stage 2 performance on both benchmarks using different features.}}
\label{tbl:feats_stage2}
\end{minipage}

\begin{minipage}{0.5\textwidth}
\makeatletter\def\@captype{table}
\centering
\scalebox{0.65}{
\begin{tabular}{l|ccc|ccc}
\hline
 & &AR@1 & & &AUC & \\
\cline{2-7}
  &TSM &I3D &TSN &TSM &I3D &TSN \\
\hline
2D-TAN  &0.2302 &0.1451 &\textbf{0.0980} &59.37 &48.69 &34.03  \\
Reloc. &0.0731 &0.0770 &0.0340 &- &- &- \\
\hline
Ours  &\textbf{0.2307} &\textbf{0.1469} &0.0956 &\textbf{61.43} &\textbf{49.76} &\textbf{35.07}\\
\hline
\end{tabular}}
\parbox{5cm}{\caption{Stage2 performance on HACS compared to reference solutions 2D-TAN \cite{zhang2020learning} and Video-Reloc. \cite{feng2018video}.}}
\label{tbl:res_stage2}
\end{minipage}
\end{minipage}
\begin{minipage}{0.5\textwidth}
\makeatletter\def\@captype{table}
\raggedright
\scalebox{0.75}{
\begin{tabular}{l|c|ccc}
\hline
Method &Config.  &Prec@1 &Prec@5 &Prec@50 \\ \hline
Trip.+2D-TAN  &HACS &0.4510 &0.4828 &0.4242 \\
Trip.+Reloc.  & &0.1645 &0.1541 &-\\
Ours &TSM &\textbf{0.4707} &\textbf{0.5060} &\textbf{0.4301}  \\ \hline

Trip.+2D-TAN &HACS &0.3109 &0.3367 &\textbf{0.2873} \\
Trip.+Reloc.   & &0.1849 &0.1637 &-\\
Ours &I3D &\textbf{0.3161} &\textbf{0.3463} &0.2783 \\ \hline

Trip.+2D-TAN &HACS &0.2756 &0.2799 &0.2378 \\
Trip.+Reloc.  & &0.1201 &0.1167 &-\\
Ours &TSN &\textbf{0.2836} &\textbf{0.2883} &\textbf{0.2469} \\ \hline

Trip.+2D-TAN &ANet &0.3084 &0.3380 &0.2843 \\
Trip.+Reloc.  & &0.2346 &0.2033 &- \\
Ours &TSM &\textbf{0.367} &\textbf{0.3577} &\textbf{0.3022}\\ \hline

Trip.+2D-TAN &ANet &\textbf{0.4752} &0.4679 &0.4539 \\
Trip.+Reloc. &  &0.1416 &0.1738 &- \\
Ours &I3D &0.4749 &\textbf{0.4688} &\textbf{0.4602} \\ \hline

Trip.+2D-TAN &ANet &0.5379 &0.5154 &\textbf{0.4079} \\
Trip.+Reloc. &  &0.1374 &0.1496 &- \\
Ours &TSN &\textbf{0.5428} &\textbf{0.5268} &0.3984 \\ \hline

\hline
\end{tabular}}
\caption{Overall SVMR performances of different methods using different features.}
\label{overall_feats}
\end{minipage}
\end{minipage}

\subsection{Comparison with Reference Solutions}
\textbf{Reference Solutions}
Based on our proposed new task, we compare our model with several reference solutions that deal with similar problems. Given ArcFace \cite{deng2019arcface} performs poorly in stage 1, we use triplet ranking model to perform stage 1 video retrieval as the reference solution. As for the stage 2, we adopt \textit{video-reloc.} \cite{feng2018video} which can localize a clip from video according to a query clip. The other reference solution is modified from 2D-TAN \cite{zhang2020learning} which is originally designed for grounding text descriptions in videos. Specifically, we replace the text query encoding branch of 2D-TAN with a video query clip encoding branch and keep the rest of 2D-TAN unchanged. In this experiment, top 10 videos are retrieved via stage 1. The stage 2 and overall experimental results are shown in Table 8
and \ref{overall_feats}, respectively. Please note that Video-Reloc. \cite{feng2018video} can only retrieve one clip per video, therefore, AUC metric and Prec@50 are missing for it in Table 8
and \ref{overall_feats}, respectively.
%
We can see our ATLN model proposed in stage 2 outperforms reference solutions from Table 8.
Its AR@1 is slightly better than 2D-TAN but the overall AUC performance is consistently better. Video-Reloc. performs much worse in this stage because it is limited to predict only one positive target clip by design and that condition is not met on our benchmark. From Table \ref{overall_feats}, we can see that our method achieves consistently better performance than these reference solutions. They validate the superiority of our framework in terms of our newly proposed SVMR task over these solutions.

\begin{figure}[t]
  \centering
  \includegraphics[width=0.9\columnwidth]{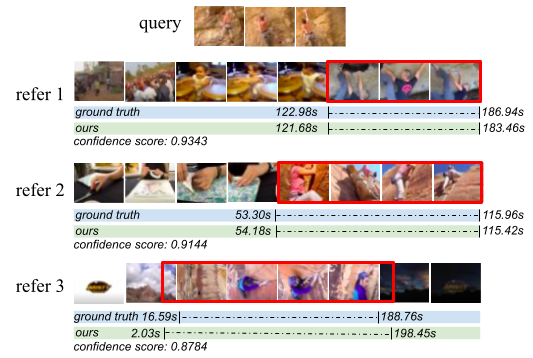}
   \caption{The top 3 retrieved clips are marked in red. Reference 1, 2 and 3 contain 3, 2 and 1 action classes, respectively. }
\label{vis}
\end{figure}

\subsection{Qualitative Results}
Figure \ref{vis} is the visualization of results of our SVMR task. The query clip is ``Rock\_climbing", and all of top 3 retrieved reference videos contain the semantic similar moments. As we can see, our proposed method can not only accurately retrieve the related candidate reference videos, but also precisely localize the corresponding segments with high confidence scores.

\section{Conclusion}
In this paper, we introduce a new task as as \textbf{S}emantically \textbf{V}ideo \textbf{M}oment \textbf{R}etrieval (SVMR) at Scale, which aims at finding corresponding videos from a large video set according to a query clip and localizing the exactly corresponding part in an untrimmed video. Besides, to address the lacking of datasets to benchmark such a new task, we adjust two large-scale datasets to make them suitable for evaluating SVMR algorithms. To address our problem, we also proposed a two-stage approach that include a two-branch auto-encoder with CTC layers to retrieve related videos, and an attention-based boundary-matching network to re-locating the precise location of semantically corresponding part in an untrimmed video. Furthermore, we conduct expensive experiments on this problem to show the remarkable performance compared to baseline methods. We hope it will inspire further research into this direction.

\bibliographystyle{bib}
\bibliography{main}
\end{document}